\documentclass[11pt]{article}

\usepackage{acl}

\usepackage{times}
\usepackage{latexsym}

\usepackage[T1]{fontenc}

\usepackage[utf8]{inputenc}

\usepackage{microtype}

\usepackage{inconsolata}

\usepackage{graphicx}

\usepackage{booktabs}
\usepackage{multirow}

\usepackage{tikz}
\usepackage{pgfplots}
\pgfplotsset{compat=1.18}
\usepackage{makecell}
\usepackage{booktabs}
\usepackage{tabularx}

\usepackage{xcolor}
\usepackage{enumitem}

\definecolor{SkyBlueLight}{HTML}{4477AA}
\definecolor{Green}{HTML}{228833}
\definecolor{DeepSpaceGreen}{HTML}{009E73}
\definecolor{Pink}{HTML}{EE6677}
\definecolor{Gray}{HTML}{BBBBBB}
\definecolor{MintWave}{HTML}{66CCEE}
\definecolor{Mustard}{HTML}{CCBB44}
\definecolor{DarkGray}{HTML}{6B6B6B}
\definecolor{Purple}{HTML}{AA3377}

\usepackage{subcaption}
\usepgfplotslibrary{colormaps}

\usepackage{longtable}
\usepackage{array}
\usepackage{ragged2e}
\newcolumntype{L}[1]{>{\RaggedRight\arraybackslash}p{#1}}

\usepackage{amsmath}

\usepackage{makecell}
\usepackage{enumitem}
\usepackage{float}

\title{You Know What I Mean: \\A Benchmark for Agentic Conversational Reference Grounding}

\author{
Karen Fuchs$^{1}$ \quad
Uri Katz$^{1}$ \quad
Yoav Goldberg$^{1,2}$ \\
$^{1}$Bar-Ilan University \qquad
$^{2}$Allen Institute for AI \\
\texttt{\{karenshakedf,urikacid,yoav.goldberg\}@gmail.com}
}

\begin{document}
\maketitle
\begin{abstract}
Collaborative conversations frequently contain references whose targets are indirect rather than named: resolving "this looks like the fix discussed yesterday" requires combining conversational context with evidence from the surrounding workspace which is accessible through APIs or user interfaces. We formalize this problem as Conversational Reference Grounding (CoRG): using a given set of tools to resolve a reference in conversation to the unique external item intended by the speaker. CoRG is challenging because it combines lexical, semantic, and temporal cues distributed across the conversation and the external workspace. Agents must translate these heterogeneous signals into effective tool use: formulating strategies, discovering plausible candidates, inspecting their metadata and content, and ruling out close alternatives. We study CoRG through \textsc{RepoRef}\footnote{\url{https://github.com/karenShaked/RepoRef-Benchmark}}, a benchmark of 400 developer-chat segments grounded in GitHub issues, pull requests, and commits across 92 repositories. Unlike single-shot retrieval tasks, \textsc{RepoRef}, often requires multi-step tool use. Our results show that CoRG remains challenging for current agents, even the best agent reaches only 67.0\% success rate, leaving one third of references unresolved. These findings position CoRG as a concrete benchmark for studying how agents search, inspect, and verify information in realistic multi-tool environments.

\end{abstract}

\section{Introduction}

We are interested in the task of grounding conversational references in digital environments. When communicating in online workspaces people often refer to external resources without naming them directly. A collaborator may write “can you update the doc from yesterday?” assuming that others can infer the intended document, ticket, post, or any other resource. Human collaborators can often resolve such references by combining prior conversational context with shared knowledge of the workspace, and by searching the available systems using the information in the conversation. We would like AI assistants operating in these environments to support the same capability:  resolving implicit conversational references to the unique external items they denote. 

\begin{figure}[t!]%
    \centering
    \includegraphics[width=\columnwidth]
    {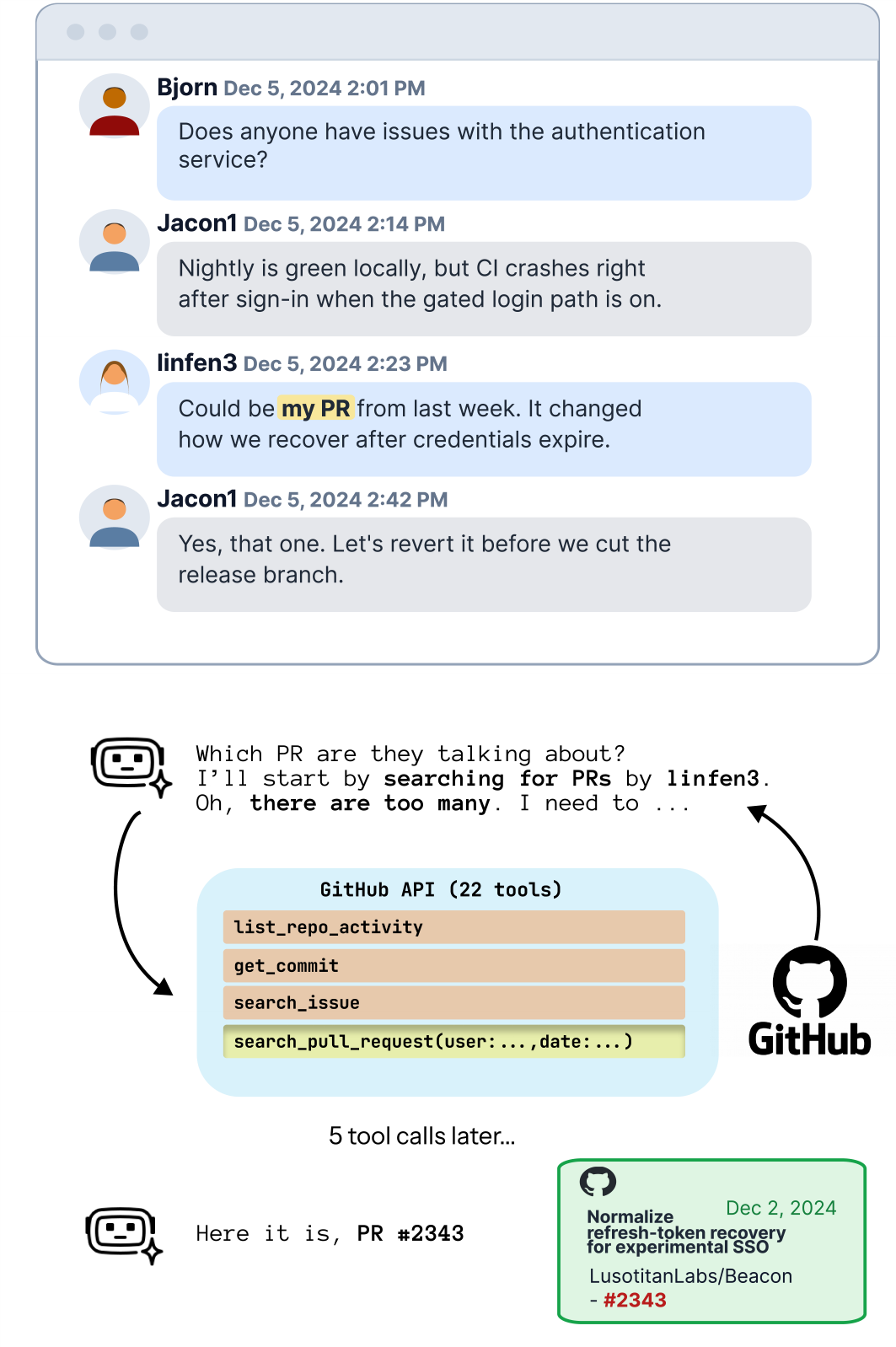}
    \caption{Illustrative example of the Conversational Reference Grounding task.}
    \label{fig:fig_label_1}
\end{figure}

We refer to this problem as Conversational Reference Grounding (CoRG). Given a multi-party conversation and an anchor message containing an underspecified reference, CoRG requires resolving that reference to the unique item in an external system intended by the speaker. 

In realistic collaborative workspaces, the relevant resources are not stored in a single static corpus. They live in external systems that change over time, and must often be accessed through platform-specific interfaces. We therefore formulate CoRG as a tool-use agent task. This follows a growing line of tool-use benchmarks that evaluate whether LLM agents can solve tasks by interacting with external APIs rather than relying only on parametric knowledge \cite{qin2023toolllmfacilitatinglargelanguage, li2023apibankcomprehensivebenchmarktoolaugmented, xie2024osworldbenchmarkingmultimodalagents,wang2026paperarenaevaluationbenchmarktoolaugmented}. 

In CoRG, reference resolution often depends on evidence that is contextually distributed or expressed without lexical overlap with the reference itself. A conversation may point to a referent through causal descriptions ("the change that caused token refresh to break"), negation ("the PR about URL helpers, but not the one Alice opened"), temporal cues ("the ticket that was opened a few minutes ago") or other mechanisms. Solving CoRG therefore requires more than retrieving topically similar artifacts; it requires reconstructing the intended reference and discriminating among close candidates. This mismatch directly challenges retrieval agents: prior work has shown that models can over-rely on surface lexical similarity, failing when relevant candidates use wording that differs from the query \cite{hagstrom-etal-2025-language,tchuindjo2026obliqbenchexposingoverlookedbottlenecks}.

 In this paper, we instantiate this setting in open-source software development: the conversations curated from Gitter channels, the external environment is GitHub, and the referents are issues, pull requests, and commits. This setting is increasingly relevant as LLM agents are evaluated and deployed in software-engineering work environments, where they must operate over repositories, issue trackers, and project-specific communication channels \cite{10.5555/3737916.3739517,xu2025theagentcompanybenchmarkingllmagents}.

We introduce \textsc{RepoRef}: a benchmark of 400 conversation segments and 7,781 messages, each containing a reference to a GitHub artifact. The benchmark covers 92 unique repositories and diverse technical code domains. The task requires an agent to interpret the conversation, search the corresponding repository, distinguish the intended target from plausible distractors, and return the exact referenced GitHub item. 

This work makes three primary contributions:
\begin{itemize}[leftmargin=*, noitemsep, topsep=2pt]
    \item \textbf{Task}. We define Conversational Reference Grounding (CoRG), where an agent must resolve a conversational reference to a unique external target using tools.
    \item \textbf{Benchmark}. We introduce \textsc{RepoRef}, a controlled benchmark that grounds real developer-chat references in GitHub issues, pull requests, and commits.
    \item \textbf{Evaluation}. We benchmark state-of-the-art LLM agents under a shared tool-use protocol, analyzing accuracy, cost, and systematic failure modes.
\end{itemize}

\section{The Conversational Reference Grounding Problem}
\subsection{Formal Definition}

We define Conversational Reference Grounding (CoRG) as follows. Given a multi-participant conversation \(C = (m_1,\ldots,m_n)\), containing an anchor message
\(m_a \in C\) referring to an underspecified resource in an external digital environment \(E\),  the task is to identify (ground) the unique external resource \(a^\star \) referred to in \(m_a\). The environment, and the resources \(A_E=\{a_1,\dots,a_k\}\) it contains, are accessible only through a set of tools (APIs) denoted by \(\mathcal{X}\): the set of resources is not accessible in other ways, and cannot be indexed locally.
The goal is to correctly identify \(a^\star\in A_E \) using the minimal number of environment interactions (tool calls).

\subsection{CoRG as a Search Task}
CoRG is an interesting and challenging agentic multi-tool search setup: to solve a CoRG instance, the agent must formulate an effective search strategy given partial information and perform a sequence of information-seeking tool calls.

Unlike standard information retrieval, which often starts with a direct question, CoRG begins with a conversation containing an indirect reference from which the agent must reconstruct an effective query strategy using distributed conversational evidence. This can be viewed as a form of "oblique query" \cite{tchuindjo2026obliqbenchexposingoverlookedbottlenecks} in an agentic setting. The initial evidence may span lexical and semantic cues, temporal and speaker information, artifact types, and artifact content and metadata. Starting from this evidence, the agent then needs to discover candidates, inspect evidence, reformulate queries, and verify competing candidates. Effective performance depends on the agent’s search strategy: how it formulates and reformulates queries, selects tools, explores candidates, and balances additional search steps against efficiency. Different instances are best served by different strategies, and a strategy may evolve during execution as new evidence is revealed.

\begin{table}[t]
\centering
\small
\begin{tabular}{lrl}
\toprule
\textbf{Value} & \textbf{Quantity} \\
\midrule
Repositories & 92 \\
Chat communities & 23 \\
Conversation segments & 400 \\
Messages & 7{,}781 \\
Unique Speakers & 532 \\
\bottomrule
\end{tabular}
\caption{Summary statistics for the RepoRef benchmark.}
\label{tab:dataset-stats}
\end{table}

\section{Benchmarking the CoRG Task}
We study CoRG in open-source software development, where references are common and externally verifiable. Developers working on a shared GitHub project often discuss issues, pull requests, commits, and repository changes in online chat channels. They sometimes point to a specific artifact by its URL or ID, and sometimes just provide an indirect reference to it. GitHub provides a concrete external environment with searchable artifacts, users, metadata, comments, diffs, and timestamps, all accessible through a set of 22 read-only GitHub API endpoints, making it a natural testbed for tool-mediated conversational grounding.

\noindent\textbf{The RepoRef benchmark. }We introduce \textsc{RepoRef},a multi-tool agentic search benchmark for resolving indirect developer-chat references to GitHub artifacts. Each \textsc{RepoRef} instance consists of a sequence of chat messages, where each message is accompanied with author id and a time stamp. 
 One of the chat messages is marked, and indirectly points to a resource (issue, commit or pull-request). The ground-truth reference resource is provided but hidden from the agent. The agent should then locate the resource using 22 provided GitHub search tools. The tools are read-only, and expose various search modalities (time, username, content, etc) following the GitHub API. The chat messages are real developer chat messages from Gitter.\footnote{Gitter is a chat platform used by open source developer communities linked to GitHub.} The agent is scored on its ability to identify the correct resource (recall@1).

\subsection{Benchmark Design Criteria}
\textsc{RepoRef} is constructed around four principal design criteria:
\begin{itemize}[leftmargin=*, noitemsep, topsep=2pt]
\item \textbf{Natural conversations.} Examples are drawn from naturally occurring conversations and real GitHub repositories, rather than generated synthetically.
\item \textbf{Identifiable references.} The intended artifact exists in the external system and is recoverable using the provided tools. 
\item \textbf{Unambiguous references.} Each conversation contains enough evidence to identify a single intended GitHub artifact.
\item \textbf{External Grounding.} The reference is indirect and cannot be resolved from the conversation alone.
\end{itemize}

\subsection{From Direct Links to Natural Implicit References}
To construct \textsc{RepoRef} we use Gitter conversations that mention a concrete GitHub artifact by ID. These establish the unique ground-truth artifacts mentioned in the conversation. We then minimally edit the referring message to remove the direct mention, and replace it with an indirect one. We then establish that the edited conversation has sufficient evidence to link the message to the correct artifact, as well as to select the correct artifact from a set of similar ones. This allows us to accommodate all the design criteria listed above.

\subsection{RepoRef Construction Pipeline}
The construction pipeline has five stages. We start from chat messages that contain links to issues, pull requests, or commits, use these links as initial labels, select the surrounding conversation needed to interpret the reference, remove the direct identifier through natural masking, and retain only cases where the intended target remains recoverable from the conversation and GitHub evidence.

\paragraph{Data Source}
We use two existing public datasets of developer communication: GitterCom \cite{parra2020gittercom} and the Gitter issue-discussion dataset \cite{sahar2021issue}. 
Together, Gitter and GitHub provide a natural CoRG setting: developers refer to GitHub records, while GitHub provides an API to metadata, comments, diffs, and timestamps. We therefore use GitHub issues, pull requests, and commits as the target referents for \textsc{RepoRef}.

\paragraph{Ground-Truth Reference Extraction}
To obtain reliable ground-truth targets at scale, we start from messages that contain direct references to GitHub issues, pull requests, or commits. We identify GitHub links using regular expressions and normalize each detected link to a canonical GitHub referent. The message containing the link is treated as the initial anchor message, and the linked GitHub record provides the ground-truth referent. This yields an externally verifiable target before masking.

\paragraph{Reference-Centered Conversation Segmentation}
\label{sec:reference-centered-segmentation}
For each anchor message, we construct a reference-centered conversation segment that aims to cover the full span of messages related to the reference. We use reference-centered segments rather than fixed-size windows because developer chat discussions are context-sensitive, entangled, and multi-participant \cite{10.1145/3412378}, so relevant evidence may appear several turns before or after the anchor message.

The backward boundary is the first message introducing the topic that leads to the reference, and the forward boundary is the last technical message responding to that topic. We include direct replies, @mentions, and messages contributing technical evidence about the same GitHub referent, following prior work that treats participant addressing and name mentions as essential cues for identifying conversational boundaries and resolving thread structure in multi-party chats \cite{elsner-charniak-2010-disentangling}. We use an LLM-assisted segmenter to apply these boundary criteria.

We evaluate the segmenter on a random sample of 50 human-labeled ideal boundaries. For each segment, human annotators marked the ideal conversation boundaries: the first message introducing the topic that leads to the reference and the last technical message responding to that topic. The segmenter matches the human boundary judgments in 96.3\% of evaluated cases. In the remaining cases, the predicted segment retains all high-importance messages but includes some extra low-relevance context, making the errors primarily precision rather than recall errors. Therefore, the conversation segments may be longer than the minimal relevant context, but are unlikely to omit evidence needed for reference resolution.

We started from 3.8k segmented conversations and retained 2.6k that contained at least two speakers and at least three messages.

\paragraph{Natural Reference Masking}
We convert each direct reference into an indirect one. A simple deletion of the URL or identifier would create unnatural text and often leave unnatural template-like traces such as “see ISSUE” or “I opened [MASK],” which do not resemble real developer communication. We therefore apply natural reference masking: the reference-bearing message is minimally rewritten, using an LLM to remove direct GitHub identifiers while preserving its pragmatic role in the conversation.

For example, a message such as “I opened https://github.com/.../issues/888 for this” may be rewritten as “I opened a ticket for this.” The rewritten message preserves the conversational act of referring to an external GitHub record, but no longer reveals its identifier, title, or URL.

To reduce leakage, masking is applied only after the ground-truth link has been extracted and normalized. To further encourage natural rewrites, the rewriting model is provided with the reference type, such as issue, pull request, or commit, together with representative examples of how that type of object is mentioned indirectly in real developer conversations. 

Its role is limited to producing a natural conversational paraphrase of only the reference anchor message; all surrounding context is kept unchanged. Upon manual inspection of 100 masked conversations, we found no conversational intent was changed.

To test whether natural masking affects task performance, we conducted an additional ablation in which the naturally rewritten references were replaced with hard-coded placeholders indicating only the artifact type, such as \texttt{<ISSUE>}. We evaluated Gemini-3-Flash on a random subset of 100 examples under both masking conditions. Performance was similar with natural masks and type-only placeholders, with success rates of 77\% and 75\%, respectively. McNemar test found no statistically significant difference between the conditions ((p = 0.83)), suggesting that the natural masking procedure does not materially alter task difficulty.

\paragraph{Ambiguity Filtering}
Direct GitHub links provide externally verifiable targets \cite{panthaplackel-etal-2022-learning}, but linked messages still require filtering to support faithful and unambiguous exact-match evaluation.

First, we verify that the conversation contains enough cues to match the reference resource. We do this by presenting the conversation and the reference to an LLM and asking it to score the chat-target match on a 0-3 scale. We retain examples with a connection score of at least 2, ensuring that the referent is grounded in the conversation. The LLM-assisted faithfulness filter showed substantial agreement with human annotation (Cohen's $\kappa = 0.77$) \cite{doi:10.1177/001316446002000104}. In the audited sample,  we did not observe unsupported retained examples, and disagreements primarily reflected conservative exclusions.

Second, we filter ambiguous examples by comparing the target referent against strong alternative candidates retrieved from the same repository. For each example, we construct an offline candidate pool from the same repository. We collect the repository's artifacts index their text and metadata locally, and retrieve top-\(k\) candidates using several broad lexical, dense, and metadata-based queries. This pool is intended to surface plausible competing GitHub records.

We then use an LLM-assisted pairwise judge to test whether it can identify the ground-truth resource over the alternative as the one matching the conversation. The judge is a strong reasoning model that is given both candidate resources, but is not given the original URL occurrence or any indication of which candidate was the linked referent. We discard an example if any alternative candidate is judged to match the conversation as well as or better than the linked referent.

This stage starts from 2.6k candidate conversations and retains 2.0k that pass the identifiability and non-ambiguity filters. The filtering is designed to keep only high-quality cases, even if this means excluding some examples that might have been valid.

\paragraph{Final benchmark Selection}
Because tool-use evaluation is costly, especially with stronger models and larger tool-call budgets, we narrow the final benchmark to 400 instances. It includes a diverse set of challenging examples targeting different solution strategies. We therefore stratify selection across seven diagnostic capability buckets assigned by operational rules over automatically computed conversation, referent, and repository features. These buckets capture stress factors such as cross-speaker evidence integration, low lexical overlap, close competing alternatives, sparse support, and commit-level targets. We thus narrow down our candidate pool of approximately 2.0k GitHub-linked conversations to 400 diverse instances of varying difficulty levels.

The final 400 instances are therefore intended as a controlled, cost-feasible, and diagnostically diverse benchmark. Full bucket definitions and counts are provided in Appendix ~\ref{app:qual}. For broader-scale evaluation, we additionally release the extended set of approximately 2,000 instances through the repository.

\begin{table*}[t]
  \centering
  \small
  \setlength{\tabcolsep}{4pt}
  \begin{tabular}{lrrrrr}
  Model & Success\% & \# Tool Calls & Avg. Total Tokens & Price per Example & Avg. Excessive Tool (n)\\
  \midrule
  Gemini-3-Flash         & \textbf{67.00} $\pm$ 2.35       & 10.51 $\pm$ 0.15       & 466.7k $\pm$ 16.6k     & 0.065\$     & 6.27(245)\\
  DeepSeek-V4-Pro        & 60.25 $\pm$ 2.45       & 8.00 $\pm$ 0.19        & 378.9k $\pm$ 33.4k     & 0.040\$    & 2.08(234)\\
  Grok-4.1-Reasoning     & 54.00 $\pm$ 2.49       & 9.56 $\pm$ 0.19        & 193.6k $\pm$ 12.0k     & 0.038\$     & 0.38(214)\\
  GLM-5.1                & 52.00 $\pm$ 2.50       & 7.55 $\pm$ 0.18        & 282.1k $\pm$ 12.3k     & 0.199\$    & 1.60(205) \\
  Claude Sonnet 4.6      & 51.25 $\pm$ 2.50       & 8.74 $\pm$ 0.26        & 191.9k $\pm$ 8.1k      & 0.317\$    & 1.80(200) \\
  GPT-5-mini             & 49.00 $\pm$ 2.50       & 6.39 $\pm$ 0.20        & 266.3k $\pm$ 17.8k     & 0.025\$   & 2.03(189)\\
  Gemini-2.5-Lite        & 4.00 $\pm$ 0.98        & 1.50 $\pm$ 0.15        & 422.5k $\pm$ 45.0k     & 0.017\$    & 1.19(16) \\
  Llama-3.3-70B          & 1.50 $\pm$ 0.61        & 2.84 $\pm$ 0.09        & 30.4k $\pm$ 4.0k       & 0.00\$    & 0.67(6) \\
  \midrule
  Claude Code Opus 4.7  & 63.25 $\pm$ 2.41 & 3.85 $\pm$ 0.19               & 252.4k $\pm$ 9.3k &  0.383\$  & 0.25(240)\\
  \bottomrule
  \end{tabular}
  \caption{Overall model performance on the CoRG benchmark at tool-call budget
  \(B=10\).}
  \label{tab:cross-model}
\end{table*}

\subsection{RepoRef Composition and Instance Format}
\textsc{RepoRef} contains 400 reference-centered conversation segments from 23 Gitter communities, comprising 7,781 messages from 532 unique speakers (Table~\ref{tab:dataset-stats}). The median conversation contains 13 messages, and every conversation includes at least two speakers. In 42\% of cases, at least one additional GitHub URL appears in the surrounding conversation context, requiring agents to disambiguate the intended reference from nearby GitHub mentions.

Each example is a sliced developer-chat conversation from a single Gitter channel, represented as timestamped messages with speaker names. One message is marked with \texttt{[REFERENCE]} and serves as the anchor: it indirectly refers to a specific GitHub issue, pull request, or commit that the model must identify using the surrounding conversation and GitHub search.

\section{Experiments}
Our experiments characterize CoRG through the behavior of current tool-using LLM agents. We aim to answer three questions: How difficult is CoRG for current agents? How does performance change with the tool-call budget? And how do agents differ in exploration strategy and failure modes?

In the primary experiment, we set a fixed max-tool budget of \(B=10\) and evaluate ReAct agents powered by a range of language models. In addition to our own ReAct implementations, we also test an agent powered by the state-of-the-art Claude-code agent harness, coupled with the state-of-the-art Claude Opus 4.7 LLM. Following the result of this experiment, we also perform an additional experiment, in which we increase the tool-call budget of the best-performing model.

\paragraph{Models.}
We evaluate seven frontier and mid-tier models, comprising a mix of open-weight and closed-source models: Claude Sonnet 4.6 \cite{anthropic2026sonnet46}, DeepSeek-V4-Pro \cite{deepseek2026v4pro}, Gemini-3-Flash-Preview \cite{google2025gemini3flash}, Gemini-2.5-Flash-Lite \cite{comanici2025gemini25pushingfrontier}, GPT-5-mini \cite{openai2026gpt5mini}, Grok-4.1-Fast-Reasoning \cite{xai2025grok41fast}, and Llama-3.3-70B \cite{ollama_llama33_70b}. All models are evaluated under a dynamic ReAct agent loop \cite{yao2023reactsynergizingreasoningacting} with a fixed tool-call budget of \(B=10\). We use each model with its provider-default settings. We additionally evaluate 
Opus 4.7 \cite{anthropic2026claudecode, anthropic2026opus47} with medium thinking effort, using the claude-code agentic harness.

\paragraph{Tools.}
The tool environment exposes search and inspection operations over GitHub issues, pull requests, commits, branches, tags, releases, labels, and repository files. Each agent is given access to 22 read-only GitHub tools, together with a special \texttt{submit\_answer} action used to return the final prediction. 

The claude-code agent is exposed to the same GitHub API, but through a read-only MCP server rather than the custom tool schemas used for the ReAct agents.

\paragraph{Metrics.}
We evaluate agents by exact-match accuracy: a prediction is correct if the submitted GitHub URL matches the gold target. We also report operational costs: average tool calls, token consumption, and estimated price per example under provider pricing with caching where applicable. To quantify search efficiency, we define \emph{Average Excessive Tool Calls}: for each example solved by at least two models, we compute how many more tool calls a correct run used than the minimal-tool-calls correct run for that example, averaged over the model's correct qualifying runs.
Additional trajectory diagnostics, including invalid tool-use, hallucinated submissions, and information-gain rates, are reported in Appendix~\ref{app:appe}.

\section{Results}

\begin{figure*}[t]
    \centering
    \includegraphics[width=0.55\linewidth]{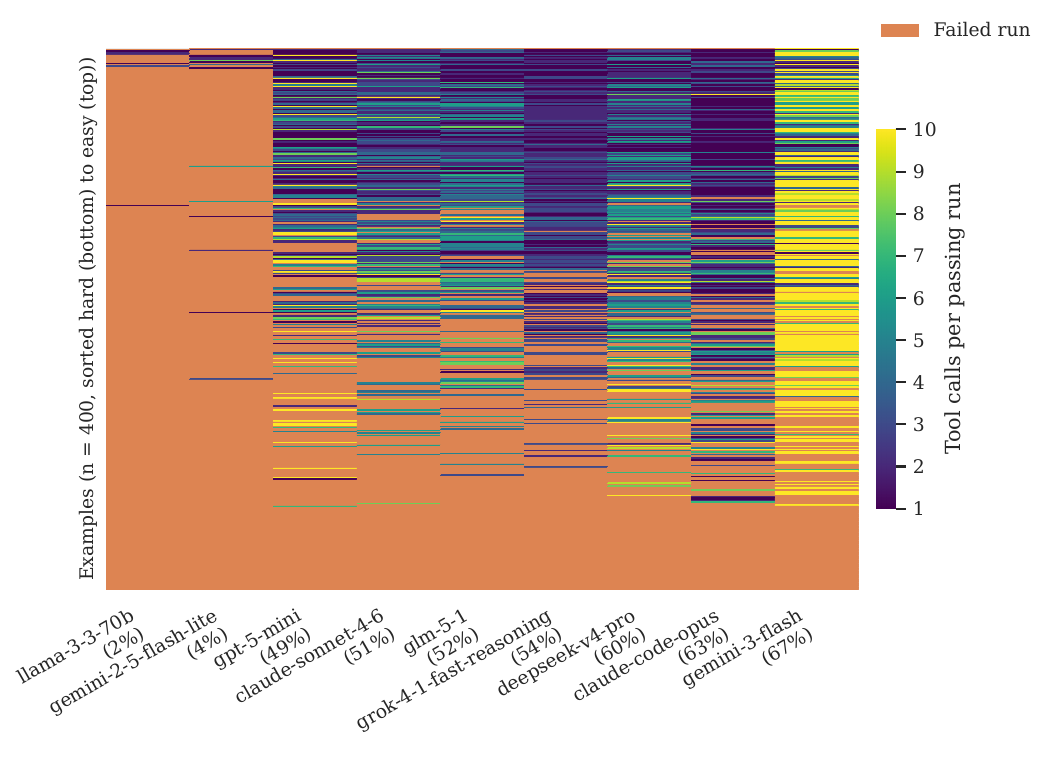}
    \caption{Per-example model outcomes on RepoRef. Each row corresponds to one benchmark instance and each column to one evaluated model;  Orange cells indicate the model failed on that instance ; For passing runs, the gradient encodes the number of tool calls used to reach the correct answer.}
    \label{fig:paper-dendrogram}
\end{figure*}

\subsection{Overall Performance: CoRG Is Challenging for Current Agents}
We first ask whether current agents can solve CoRG reliably under a fixed tool-use budget. The results are summarized in Table \ref{tab:cross-model}. While the agents achieve varying degrees of success, they are all far from perfect performance. The best performing ReAct model, Gemini-3-Flash, reaches 67.0\% accuracy on the 400-example benchmark. DeepSeek-V4-Pro follows at 60.25\%, while the remaining strong agents fall between 49.0\% and 54.0\%. Lightweight models such as Llama-3.3-70B and Gemini-2.5-Flash-Lite perform much worse, reaching success rate of 1.5\% and 4.0\% respectively. 
Claude Code with Opus 4.7, using Anthropic's agentic claude-code harness, reaches 63.25\%, ranking below the 67.0\% of the best performing Gemini-3-Flash based ReAct agent, and above the other ReAct agents.

Gemini-2.5-Flash-Lite does not issue any tool call in 65.5\% of examples, while Llama-3.3-70B has a 72.9\% tool-error rate as shown in Table~\ref{tab:tool-call-counts-error-rate}.   

Generally speaking, the success rates are not correlated with token costs, with the most expensive models, GLM-5.1 and Claude Sonnet 4.6, ranking middle-low.  

The high performance of the Gemini-3-Flash model comes at a cost: it uses many more tool calls than required, with Average Excessive Tool value of 6.27. 

\subsection{Per-instance Difficulty}
Figure~\ref{fig:paper-dendrogram} shows per-instance success and failure patterns across all models. Each column is a model, and each row represents a test instances. The colors indicate both success or failure to solve an instance, as well as the number of tool calls used in successful solutions.

Some instances (at the bottom) are challenging for all models. These account for 18.5\% of the instances. To verify that these universally failed instances were not construction errors, we manually inspected 50 examples that were not solved by any model. In all inspected cases, the conversation and the available GitHub evidence still supported the gold artifact, suggesting that these failures reflect genuine task difficulty rather than invalid or unresolvable examples. Among the instances solved by at least one model, there is no single ordering of difficulty. Different models succeed on different subsets of examples, rather than stronger models consistently solving all examples handled by weaker ones.

Moreover, there is no clear association between instances and the number of tool calls required to solve them. The variance of the tool calls required to solve an instance is high. The most efficient model, Grok-4.1-fast-reasoning, solves many instances with only 3 tool calls or fewer. However, this comes at a price of failing to resolve a high number of instances.

\subsection{Tool Budget Improves CoRG Performance on Gemini-3-Flash}
To measure how increasing the tool-call budget affects CoRG performance, we run a sweep on a 280-example subset, covering 70\% of the full benchmark. Increasing the tool-call budget substantially improves Gemini-3-Flash performance, from 23.21\% at \(B=1\) to 73.93\% at \(B=16\). Most gains occur by \(B=6\), where accuracy reaches 63.93\%, suggesting that a moderate amount of exploration is sufficient for many examples. However, the additional gains come at substantial tool and token cost. These results are specific to Gemini-3-Flash, whose broader exploration behavior distinguishes it from the other evaluated ReAct agents; we analyze these cross-model exploration differences in Section~\ref{tool_eff}.

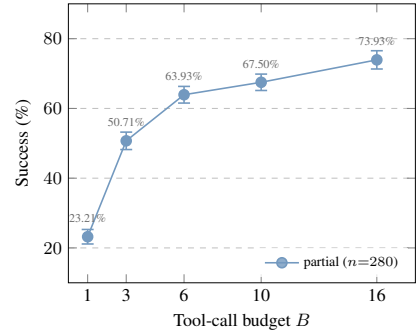
\begin{figure}[t]
    \centering
    \small
    \scalebox{0.75}{
    \begin{tikzpicture}
    \begin{axis}[
        width=\linewidth,
        height=6.5cm,
        xlabel={Tool-call budget $B$},
        ylabel={Success (\%)},
        ymin=10, ymax=90,
        xmin=0, xmax=18,
        xtick={1,3,6,10,16},
        xticklabels={1,3,6,10,16},
        x tick label style={font=\small},
        y tick label style={font=\small},
        ymajorgrids=true,
        grid style=dashed,
        legend style={
            at={(0.98,0.02)},
            anchor=south east,
            font=\scriptsize,
            draw=none, fill=white, fill opacity=0.8, text opacity=1,
            cells={anchor=west}
        },
    ]

    \addplot[
        color=SkyBlueLight!75,
        thick,
        mark=*, mark size=2.5pt, mark options={fill=SkyBlueLight!75},
        error bars/.cd, y dir=both, y explicit,
            error bar style={SkyBlueLight!75, line width=0.8pt},
            error mark options={SkyBlueLight!75, line width=0.8pt, rotate=90, mark size=3pt}
    ] coordinates {
        (1, 23.21) +- (0, 2.08)
        (3, 50.71) +- (0, 2.50)
        (6, 63.93) +- (0, 2.39)
        (10, 67.50) +- (0, 2.35)
        (16, 73.93) +- (0, 2.62)
    };

    \addlegendentry{partial ($n{=}280$)}

    \node[above=4pt, font=\tiny, color=DarkGray!60!DarkGray] at (axis cs:1,  23.21) {23.21\%};
    \node[above=4pt, font=\tiny, color=DarkGray!60!DarkGray] at (axis cs:3,  50.71) {50.71\%};
    \node[above=4pt, font=\tiny, color=DarkGray!60!DarkGray] at (axis cs:6,  63.93) {63.93\%};
    \node[above=4pt, font=\tiny, color=DarkGray!60!DarkGray] at (axis cs:10, 67.50) {67.50\%};
    \node[above=4pt, font=\tiny, color=DarkGray!60!DarkGray] at (axis cs:16, 73.93) {73.93\%};

    \end{axis}
    \end{tikzpicture}
    }
    \caption{Gemini-3-Flash accuracy as a function of tool-call budget $B$ on the 280-example budget-sweep subset.}
    \label{fig:gemini-budget-sweep}
\end{figure}

\section{Analysis}
The results above show that CoRG is challenging and that additional tool use can improve performance. We now analyze agent trajectories to understand where the challenges come from.

\subsection{Exploration Behavior and Tool-Call Efficiency}
\label{tool_eff}
To characterize exploration behavior, we analyze whether agents surface the gold target, how discovery changes across tool-call steps, and how efficiently agents stop once they have enough evidence. We first observe that CoRG performance is tied to exploration: when agents surface the gold artifact, they often select it correctly. To better localize failures, we distinguish between \emph{pre-surfacing failures}, where the gold artifact is never discovered, and \emph{post-surfacing failures}, where it is surfaced but not selected. Among stronger models, 70--92\% of failures occur before surfacing, while the gold is selected correctly in 87--91\% of cases once surfaced. A full model-wise decomposition is provided in Appendix~G. Thus, many CoRG failures occur before the final decision step: agents fail to surface the correct artifact in the first place.

The trajectory curves in Figure~\ref{fig:discovery-latency} show that models differ in how they explore in CoRG settings. Gemini-3-Flash continues to surface gold targets later in the trajectory, reaching 75.0\% gold discovery by step 10. Claude Code Opus 4.7 also surfaces the gold target in 72.0\% of examples. Several other agents plateau after only a few tool calls, suggesting that they stop exploring or fail to reformulate search effectively after the first few tool calls. This helps explain Gemini-3-Flash's stronger overall performance: broader exploration increases the chance that the gold artifact enters the candidate set.

However, high candidate recall can be achieved with different levels of tool-call efficiency. Gemini-3-Flash achieves the highest accuracy, but it also uses the most excess tool calls, with an average of 6.27 excess calls among solved examples as specified in Table \ref{tab:cross-model}. By contrast, Grok-4.1 is the most parsimonious ReAct agent when correct, with only 0.38 excess tool calls on average among solved examples. In the end-to-end setting, Claude Code Opus 4.7 is even more tool-efficient, with an average of 0.25 excess tool calls. 

The comparison across models suggests that Gemini-3-Flash follows a different exploration strategy from the other evaluated agents. Since most strong models reliably select the gold artifact once it is surfaced, this suggests that Gemini’s advantage stems primarily from broader and more persistent exploration, which increases the probability of discovering the correct candidate. Claude Code Opus 4.7 illustrates the opposite end of this trade-off: it is considerably more tool-efficient, but its more conservative exploration results in slightly lower overall accuracy. Strong performance requires both surfacing the correct artifact and stopping once enough evidence has been gathered to be efficient.

\begin{figure}[t]
    \centering
    \small
    \scalebox{0.85}{
    \begin{tikzpicture}
    \begin{axis}[
        width=\columnwidth,
        height=4.8cm,
        xlabel={Tool-call step $k$},
        ylabel={GT discovered by step $k$ (\%)},
        xmin=1,
        xmax=10,
        ymin=0,
        ymax=80,
        xtick={1,2,3,4,5,6,7,8,9,10},
        ymajorgrids=true,
        grid style={solid, gray!18},
        line width=1pt,
        no markers,
        tick label style={font=\scriptsize},
        label style={font=\scriptsize},
        axis line style={gray!45},
        tick style={gray!45},
        legend style={
            at={(0.5,-0.28)},
            anchor=north,
            legend columns=2,
            draw=none,
            fill=none,
            font=\tiny,
            /tikz/every even column/.append style={column sep=0.12cm}
        },
        legend cell align={left},
    ]

    \addplot[color=MintWave, smooth] coordinates {
        (1,24.5) (2,42.5) (3,52.5) (4,59.5) (5,62.5)
        (6,66.0) (7,69.2) (8,72.0) (9,73.5) (10,75.0)
    };

    \addplot[color=Green, smooth] coordinates {
        (1,22.0) (2,32.0) (3,37.8) (4,38.5) (5,38.5)
        (6,38.5) (7,38.5) (8,38.5) (9,38.5) (10,38.5)
    };

    \addplot[color=SkyBlueLight, smooth] coordinates {
        (1,21.8) (2,31.5) (3,39.2) (4,44.0) (5,47.2)
        (6,49.0) (7,51.8) (8,52.2) (9,53.0) (10,53.8)
    };

    \addplot[color=Mustard, smooth] coordinates {
        (1,21.0) (2,32.2) (3,39.2) (4,42.8) (5,44.8)
        (6,47.0) (7,48.5) (8,49.0) (9,49.2) (10,49.2)
    };

    \addplot[color=DarkGray, smooth] coordinates {
        (1,18.5) (2,30.2) (3,38.2) (4,45.0) (5,50.5)
        (6,53.5) (7,54.8) (8,55.2) (9,55.5) (10,55.8)
    };

    \addplot[color=Pink, smooth] coordinates {
        (1,12.8) (2,20.8) (3,27.0) (4,34.5) (5,40.8)
        (6,43.2) (7,44.2) (8,44.8) (9,45.0) (10,45.0)
    };

    \addplot[color=black, smooth] coordinates {
        (1,2.2) (2,3.5) (3,3.8) (4,4.2) (5,4.2)
        (6,5.0) (7,5.0) (8,5.2) (9,5.2) (10,5.2)
    };

    \addplot[color=Purple, smooth] coordinates {
        (1,0.2) (2,0.8) (3,0.8) (4,0.8) (5,0.8)
        (6,0.8) (7,0.8) (8,0.8) (9,0.8) (10,0.8)
    };

    \addplot[color=blue, smooth] coordinates {
        (1,32.7) (2,49.4) (3,56.4) (4,62.7) (5,65.7)
        (6,68.5) (7,69.8) (8,71.0) (9,71.8) (10,72.0)
    };

    \legend{
        Gemini-3-Flash,
        Grok-4.1-Fast,
        GPT-5-mini,
        Claude Sonnet 4.6,
        DeepSeek-V4-Pro,
        GLM-5.1,
        Gemini-2.5-Lite,
        Llama-3.3-70B,
        Claude-Code Opus 4.7
    }

    \end{axis}
    \end{tikzpicture}
    }
    \caption{Discovery latency per model over 400 capability examples. Each curve shows the percentage of runs where GT appeared by step $k$.}
    \label{fig:discovery-latency}
\end{figure}
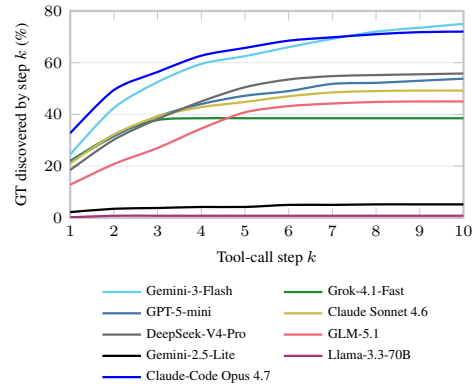

\begin{figure*}[t]
    \centering
    \small
    \scalebox{0.85}{
    \begin{tikzpicture}
    \begin{axis}[
    ybar,
    bar width=4pt,
    width=\textwidth,
    height=5cm,
    ymin=0,
    ymax=90,
    ylabel={Success (\%)},
    symbolic x coords={SurfaceMismatch,Competitive,SparseEvidence,CommitLevel},
    xtick=data,
    xticklabels={
        \shortstack{Surface\\mismatch},
        \shortstack{Competitive\\alternative},
        \shortstack{Sparse\\evidence},
        \shortstack{Commit-level\\grounding}
    },
    x tick label style={
        font=\scriptsize,
        align=center,
        text width=1.8cm
    },
    ymajorgrids=true,
    grid style=dashed,
    enlarge x limits=0.10,
    cycle list={
        {fill=SkyBlueLight, draw=SkyBlueLight},
        {fill=Pink, draw=Pink},
        {fill=Gray, draw=Gray},
        {fill=Mustard, draw=Mustard},
        {fill=MintWave, draw=MintWave},
        {fill=DarkGray, draw=DarkGray},
        {fill=Purple, draw=Purple},
        {fill=Green, draw=Green},
        {fill=DeepSpaceGreen, draw=DeepSpaceGreen},
    },
    legend image code/.code={
    \draw[#1] (0cm,-0.08cm) rectangle (0.18cm,0.08cm);
    },
    legend style={
        at={(0.5,-0.22)},
        anchor=north,
        legend columns=3,
        draw=none,
        fill=none,
        font=\scriptsize,
        /tikz/every even column/.append style={column sep=0.25cm}
    },
    legend cell align={left},
]

    \addplot coordinates { (SurfaceMismatch,77.94) (Competitive,63.64) (SparseEvidence,71.01) (CommitLevel,21.28)};
    \addplot coordinates { (SurfaceMismatch,76.47) (Competitive,43.64) (SparseEvidence,65.22) (CommitLevel,46.81)};
    \addplot coordinates { (SurfaceMismatch,69.12) (Competitive,56.36) (SparseEvidence,63.77) (CommitLevel,14.89)};
    \addplot coordinates { (SurfaceMismatch,63.24) (Competitive,43.64) (SparseEvidence,60.87) (CommitLevel,12.77)};
    \addplot coordinates { (SurfaceMismatch,57.35) (Competitive,47.27) (SparseEvidence,63.77) (CommitLevel,12.77)};
    \addplot coordinates { (SurfaceMismatch,57.35) (Competitive,52.73) (SparseEvidence,59.42) (CommitLevel,8.51)};
    \addplot coordinates { (SurfaceMismatch,52.94) (Competitive,38.18) (SparseEvidence,59.42) (CommitLevel,19.15)};

    \legend{
        Gemini-3-Flash,
        Claude Code Opus 4.7,
        DeepSeek-V4-Pro,
        Grok-4.1-Reasoning,
        GLM-5.1,
        Claude Sonnet 4.6,
        GPT-5-mini
    }

    \end{axis}
    \end{tikzpicture}
    }
    \caption{Model success rate on diagnostic capability buckets. Buckets capture surface mismatch, competitive alternatives, sparse conversational evidence, and commit-level grounding.}
    \label{fig:four-capability-bucket}
\end{figure*}
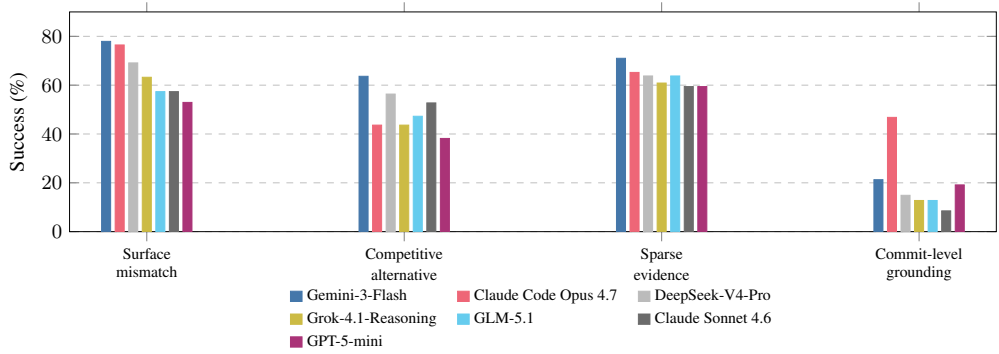

\subsection{What Makes Conversational Grounding Difficult?}

We group instances by potential sources of difficulty and report model behavior on each subgroup. The definitions and counts for all seven groups are provided in Appendix~\ref{app:qual}; here, we focus on four representative groups - surface mismatch, competitive alternatives, sparse evidence, and commit-level grounding- that illustrate the main trends.

Figure~\ref{fig:four-capability-bucket} shows that all four phenomena pose difficulty for current agents. Even in surface-mismatch cases, where the main challenge is bridging different wording between the conversation and the target artifact, no model exceeds 80\% success and several models fall near or below 60\%. Thus, lexical mismatch is already a non-trivial source of error. Sparse-evidence cases show a similar pattern: models can sometimes recover from limited conversational support, but success rates remain far from saturated.

The harder cases involve ambiguity and artifact-specific verification. Performance drops with competitive alternatives, where another artifact matches some conversational cues but is not the intended referent. The largest performance drop appears in commit-level grounding: across models, references to commits are substantially harder to ground than references to issues or pull requests. This likely reflects a combination of factors: commits are less textually salient, are harder to search for directly, and often require more precise verification against code changes or short messages.

Overall, the analysis suggests that CoRG is difficult even under familiar semantic-matching challenges, but becomes especially difficult when agents must search the correct artifact space, compare plausible candidates, and verify the intended referent.

\section{Related Work}

\paragraph{Conversational retrieval and entity linking.}
Prior work studies context-dependent and underspecified language in dialogue, including conversational grounding failures \citep{shaikh-etal-2025-navigating}, conversational retrieval and query rewriting \citep{dalton2020cast,anantha-etal-2021-open,wu-etal-2022-conqrr}, and conversational entity linking \citep{10.1145/3404835.3463258,hoveyda-etal-2024-real}. Unlike settings with an explicit information-seeking turn or predefined retrieval target, CoRG requires the system to infer the intended external artifact from distributed conversational cues, even when the speaker does not issue a query or name the artifact. POSR is closest in spirit, retrieving external reference materials from conversation segments \citep{wang-etal-2024-problem}; CoRG studies this idea in operational environments, where targets are concrete artifacts requiring tool-mediated search and disambiguation.

\paragraph{Repository-level and software-agent benchmarks.}
Recent software-engineering benchmarks evaluate repository-level code retrieval and completion \citep{liu2024repobench}, issue resolution and bug fixing \citep{zhang-etal-2023-repocoder,jimenez2024swe,10.1145/3715754}, and agentic workflows for editing code, running tests, or refining requirements \citep{10.5555/3737916.3739517,wang2025openhands,10.5555/3780338.3782245,kuang2026reagent}. Our work adds a preceding grounding step: resolving informal developer references to the concrete repository artifacts during conversations.

\paragraph{Tool-use and workplace agents.}
Benchmarks for tool-use and workspace agents evaluate how language models reason and act over APIs, web environments, and simulated workplaces \citep{10.1145/3704435,zhou2024webarena,mialon2024gaia,xu2025theagentcompanybenchmarkingllmagents}. \textsc{RepoRef} studies this capability in repository navigation through the GitHub API, but grounds it in conversational context: agents must translate indirect conversational clues into tool-mediated search and verification. This positions CoRG between conversational grounding and tool-use evaluation, where success depends on both grounding accuracy and the agent's search strategy.

\section{Conclusion}
We introduced Conversational Reference Grounding (CoRG), a tool-mediated search task in which an agent must resolve an indirect conversational reference to a unique external item. We presented \textsc{RepoRef}, a benchmark of developer-chat segments grounded in real GitHub issues, pull requests, and commits. CoRG captures a common but underexplored requirement for agents in collaborative workspaces: converting conversational evidence into targeted search, inspection, and verification over external systems using tools. 

Our evaluation shows that this capability is still lacking in current agents. Failures often arise not only from final selection errors, but from ineffective exploration: agents fail to surface, inspect, or verify the right artifact. More broadly, \textsc{RepoRef} provides a concrete setting for studying how agents use tools to operationalize conversational context. The results point to higher-recall exploration, better use of metadata and temporal cues, and lightweight candidate verification as promising directions for improving performance in this task.

\section*{Limitations}
Our benchmark is constructed from naturally occurring conversations by minimally rewriting messages that originally contained direct GitHub links, replacing the  identifier with an indirect reference. This construction raises a potential concern: conversations that originally contained direct links may differ systematically from conversations in which speakers naturally refer to an artifact indirectly. 

However, several aspects of our design mitigate this concern. First, we retain only conversations in which the intended artifact remains verifiable from the surrounding conversational and GitHub evidence after the link is removed. Second, our goal is not limited to modeling cases in which the referent is explicitly and thoroughly described. CoRG is intended to capture settings where collaborators rely on shared history and communicate through sparse, distributed cues that are sufficient for participants familiar with the context, but challenging for an external agent to interpret. For these reasons, together with our manual inspection of the constructed examples, we believe that the resulting task provides a useful proxy for naturally occurring conversational reference grounding. RepoRef is limited to open-source GitHub-based collaboration and to issues, pull requests, and commits, so it may not capture reference patterns or target types found in other digital workspaces.

Our evaluation uses a fixed read-only tool environment and limited tool budget, so results may vary under different tools, retrieval systems, budgets, or agent configurations.

\section*{Ethical considerations}
Our benchmark is constructed from datasets of Gitter channels associated with open-source software communities. Public Gitter messages are licensed under Creative Commons BY-NC-SA, and the user-generated data is publicly accessible. We use the data solely for non-commercial research purposes. We do not conduct analyses intended to identify individual users, profile them, or link messages to external personal information. We release the dataset under a Creative Commons BY-NC-SA license.
\bibliography{custom}
\appendix
\onecolumn
\section{Qualitative Examples}
\label{app:qual}
\begingroup
\small
\setlength{\parindent}{0pt}
\newcommand{\refmark}{%
  \colorbox{yellow!20}{\strut\textbf{REFERENCE}}%
}

\newcommand{\bucketheader}[2]{%
  \vspace{10pt}
  \noindent
  \colorbox{gray!12}{%
    \parbox{\dimexpr\textwidth-2\fboxsep\relax}{%
      \textbf{#1}\hspace{0.8em}\textbf{#2}
    }%
  }
  \vspace{7pt}
}

\newcommand{\subhead}[1]{%
  \vspace{5pt}
  {\textbf{#1}}\par
  \vspace{2pt}
}

\noindent
The examples below illustrate the seven diagnostic capability buckets used in RepoRef.
Each example is drawn from the \emph{hard} subset and highlights a distinct source of difficulty in conversational reference grounding.
Speakers are anonymized.

\bucketheader{C1}{Cross-speaker evidence integration}

\subhead{Conversation}

\textbf{[user\_l]} the exact same app with the exact same configuration works well on safari\\
\textbf{[user\_l]} but the same app packaged for android works perfectly there\\
\ldots\\
\textbf{[user\_a]} hmm it turns out that on iOS cordova uses UIWebView instead of WKWebView\\
\textbf{[user\_a]} WKWebView is the one that safari uses, but because of a serious bug cordova still uses UIWebView, so the performance issue could be caused by UIWebView\\
\ldots\\
\textbf{[user\_r]} 1.3 upgrade was surprisingly painless. As long as you don't replace too many meteor packages with npm packages it should be fine\\
\textbf{[user\_r]} [\refmark] Particularly described in a GitHub issue\\
\ldots

\subhead{Ground-truth artifact}

\textbf{Title:}
\textit{Meteor 1.3 beta (modules, mobile, and testing)}

\medskip
\textbf{GitHub evidence:}
Meteor 1.3 upgrades Cordova dependencies, uses \texttt{WKWebView} on iOS for improved JavaScript performance, and includes rewritten plugins for iOS and Android.

\medskip
\textbf{Conversation clue:}
Participants diagnose an iOS Cordova performance issue, contrast Safari/iOS/Android behavior, identify \texttt{UIWebView} vs.\ \texttt{WKWebView}, and point to Meteor 1.3 as the likely fix path.

\subhead{Why this is challenging}

\begin{itemize}[leftmargin=1.4em, nosep]
    \item Evidence comes from at least four participants.
    \item Relevant clues are distributed across multiple turns.
    \item No single message uniquely identifies the target.
\end{itemize}

\bucketheader{C2}{Surface mismatch}

\subhead{Conversation}

\textbf{[user\_l]} I have. Could not spot any significant changes -- but there probably is. BTW, I was not able to build the v2 branch. Got a lot of errors (Node 5.3.0)\\
\textbf{[user\_a]} [\refmark] yes, there were some errors I fixed yesterday in the pull request I opened. Now, the master branch should be error free again\\
\textbf{[user\_l]} @user\_a Thank you for taking the time to look at the code. I think that I, for now, stick with 1.x component pattern.

\subhead{Ground-truth artifact}

\textbf{Title:}
\textit{fixing eslint related errors and warnings}

\medskip
\textbf{GitHub evidence:}
A code-quality PR addressing ESLint-flagged errors and warnings on the v2 branch of \texttt{material-design-lite}.

\medskip
\textbf{Conversation clue:}
The chat refers only to ``errors'' and the ``master branch.'' It does not mention \texttt{eslint}, \texttt{warnings}, or other distinctive vocabulary that directly surfaces the PR.

\subhead{Why this is challenging}

\begin{itemize}[leftmargin=1.4em, nosep]
    \item Low lexical overlap between chat and target.
    \item Many similarly worded artifacts exist in the repository.
    \item Surface-level keyword matching is insufficient.
\end{itemize}

\bucketheader{C3}{Author-crowded disambiguation}

\subhead{Conversation}

\textbf{[user\_m]} Any idea when this issue will be fixed? \url{https://github.com/MonoGame/MonoGame/issues/6045}\\
\textbf{[user\_m]} Could be an easy fix, will bring love and money\\
\textbf{[user\_r]} @user\_m it's harder than it looks, but mostly just an api swap needed to sort it out without causing new issues\\
\textbf{[user\_c]} I am unable to reproduce the issue\\
\textbf{[user\_c]} oh no, wait \ldots\ are you talking about the same frame? I just noticed it does update in the next frame\\
\textbf{[user\_c]} [\refmark] @user\_m I just put up a PR for that\\
\textbf{[user\_m]} Woah, @user\_c merci!

\subhead{Ground-truth artifact}

\textbf{Title:}
\textit{[SDL] Optimize mouse position tracking (fixes \#6045)}

\medskip
\textbf{GitHub evidence:}
An SDL-backend PR that fixes the same-frame mouse-position-tracking bug in issue \#6045 by tracking position via the motion event when the window has mouse focus.

\medskip
\textbf{Conversation clue:}
\texttt{user\_c} reproduces the same-frame issue and then says, ``I just put up a PR for that.'' The agent must identify which of \texttt{user\_c}'s many recent MonoGame PRs is intended.

\subhead{Why this is challenging}

\begin{itemize}[leftmargin=1.4em, nosep]
    \item The target author has a large artifact history.
    \item Several artifacts are close in time to the conversation.
    \item Resolving the reference requires combining author and temporal cues.
\end{itemize}

\bucketheader{C4}{Competitive alternative}

\subhead{Conversation}

\textbf{[user\_t]} One thing I would love from MonoGame was splitting out all the math stuff into a separate project. Feels annoying that I have to reference DirectX or OpenGL frameworks on my headless server.\\
\textbf{[user\_v]} [\refmark] @user\_t I believe that's been discussed in an issue already\\
\textbf{[user\_t]} Ah, that looks good. And yeah, was exactly what I had in mind. Basically separate all the platform independent stuff.\\
\textbf{[user\_v]} I've been asking that for years\ldots\ thing is, monogame development moves reaaallly slowly, due to backwards compatibility and console support.

\subhead{Ground-truth artifact}

\textbf{Title:}
\textit{Extract platform-agnostic classes from MG.Framework into separate assembly}

\medskip
\textbf{GitHub evidence:}
The issue proposes extracting platform-agnostic classes such as \texttt{Math} and \texttt{Vector} into a separate \texttt{MonoGame.Core} assembly for headless and non-graphical use. It explicitly states that it restarts discussion from issue \#2500.

\medskip
\textbf{Conversation clue:}
\texttt{user\_t} requests exactly such a split for headless server use, and \texttt{user\_v} points to ``an issue already.''

\medskip
\textbf{Competing candidate:}
Issue \#2500 and other earlier discussions about platform-agnostic separation are also plausible referents.

\subhead{Why this is challenging}

\begin{itemize}[leftmargin=1.4em, nosep]
    \item The correct artifact and competing artifacts are both plausible referents.
    \item They are similar enough to be confused.
    \item They still differ in details that allow the correct answer to be recovered.
\end{itemize}

\bucketheader{C5}{No author shortcut}

\subhead{Conversation}

\textbf{[user\_s]} BTW, What's the Amber equivalent of \texttt{root\_url} or any other \texttt{\_url} methods from Rails? Something which takes into account the requesting domain?\\
\textbf{[user\_e]} At the moment we do not have url helpers we started\ldots\\
\textbf{[user\_s]} @user\_e Is there a link to the discussion? I'd like to help\\
\textbf{[user\_e]} give me a sec --- I swear @[user\_f] opened an issue for this\\
\textbf{[user\_e]} [\refmark] @user\_s you can post your thoughts on the issue I found

\subhead{Ground-truth artifact}

\textbf{Title:}
\textit{automatically create paths helpers}

\medskip
\textbf{GitHub evidence:}
A feature-request issue proposing automatically generated path and URL helpers similar to Rails.

\medskip
\textbf{Conversation clue:}
\texttt{user\_s} asks about Rails-style URL helpers in Amber. \texttt{user\_e} retrieves an issue opened by \texttt{user\_f}. The ground-truth author is mentioned in the conversation but is not the speaker making the masked reference.

\subhead{Why this is challenging}

\begin{itemize}[leftmargin=1.4em, nosep]
    \item The speaker making the reference is not the artifact author.
    \item A simple speaker-to-author shortcut would fail.
    \item The agent must track conversational roles correctly.
\end{itemize}

\bucketheader{C6}{Sparse conversational evidence}

\subhead{Conversation}

\textbf{[user\_w]} If someone wants an ``official'' response to something from a Microsoft employee they should open a support case shouldn't they? or does corefx come with an implied warranty of some kind?\\
\textbf{[user\_j]} LICENSE.TXT is MIT \ldots\ THE SOFTWARE IS PROVIDED ``AS IS'', WITHOUT WARRANTY OF ANY KIND\\
\textbf{[user\_w]} [\refmark] that's what I thought, dude seems to be treating it as a support channel which it isn't, based on that issue he opened\\
\textbf{[user\_j]} He came into the conversation somewhat upset. If the team shows him that they care about fixing the problem he found, it will go a long way\ldots

\subhead{Ground-truth artifact}

\textbf{Title:}
\textit{SqlClient. Inappropriate behavior after a commit timeout}

\medskip
\textbf{GitHub evidence:}
A SqlClient bug report about commit-timeout handling, reproduced across both \texttt{System.Data.SqlClient} and \texttt{Microsoft.Data.SqlClient} on .NET Core 2.2 and .NET Framework 4.8.

\medskip
\textbf{Conversation clue:}
The reference appears only as a passing remark in a broader meta-discussion about expectations of Microsoft support in open-source channels. The actual SqlClient issue is not described; the primary anchor is ``that issue he opened.''

\subhead{Why this is challenging}

\begin{itemize}[leftmargin=1.4em, nosep]
    \item The target is supported by only a small amount of conversational evidence.
    \item There is no repeated or extended discussion of the target.
    \item The link is therefore weaker than in cases with deeper multi-message engagement.
\end{itemize}

\bucketheader{C7}{Commit-level grounding}

\subhead{Conversation}

\textbf{[user\_m]} there is also a yummy food photo showing up in the app. not sure, where it comes from\\
\textbf{[user\_d]} @user\_m That image comes from here \texttt{sample/FOSSASIA16/tracks\#L526}\\
\textbf{[user\_m]} do you know where that images comes from and who added it?\\
\textbf{[user\_n]} Just a sec. I'll add the commit number\ldots\ Okay not getting the exact commit where this is done. GitHub is not showing it\\
\ldots\\
\textbf{[user\_n]} Okay here it is\\
\textbf{[user\_n]} [\refmark] I found the commit where it was added.\\
\textbf{[user\_n]} @user\_m @user\_s added this

\subhead{Ground-truth artifact}

\textbf{Title:}
\textit{Update sample track metadata}

\medskip
\textbf{GitHub evidence:}
A small commit to the \texttt{open-event} sample data that adds or edits the \texttt{tracks} JSON entry introducing the \texttt{lorempixel.com/400/200/} placeholder image URL referenced in the conversation.

\medskip
\textbf{Conversation clue:}
Participants trace the unexpected ``yummy food photo'' through \texttt{tracks\#L526} and the file history until one participant announces that they found the commit where it was introduced.

\subhead{Why this is challenging}

\begin{itemize}[leftmargin=1.4em, nosep]
    \item The target is a commit rather than an issue or PR.
    \item Resolution requires tracing file history.
    \item The target contains relatively little descriptive text.
\end{itemize}

\endgroup
\clearpage

\section{Diagnostic Buckets Performance}

We report model performance across seven diagnostic buckets, each capturing a different source of difficulty in conversational reference grounding: cross-speaker evidence integration, surface mismatch, author-crowded disambiguation, competitive alternatives, absence of author shortcuts, sparse conversational evidence, and commit-level grounding. Table~\ref{tab:per-capability-bucket} shows accuracy within each bucket. 

\begin{table}[H]
\centering
\small
\caption{Model accuracy (\%) across diagnostic buckets.}
\label{tab:per-capability-bucket}
\setlength{\tabcolsep}{3pt}

\begin{tabular}{lrrrrrrr}
\toprule
\textbf{Model} &
\textbf{\shortstack{Cross-speaker\\evidence}} &
\textbf{\shortstack{Surface\\mismatch}} &
\textbf{\shortstack{Author-crowded\\disambig.}} &
\textbf{\shortstack{Competitive\\alternative}} &
\textbf{\shortstack{No author\\shortcut}} &
\textbf{\shortstack{Sparse\\evidence}} &
\textbf{\shortstack{Commit-level\\grounding}} \\
\midrule

Gemini-3-Flash
& \textbf{69.39}
& \textbf{77.94}
& \textbf{80.70}
& \textbf{63.64}
& \textbf{74.55}
& \textbf{71.01}
& 21.28 \\

Claude Code Opus 4.7
& 61.22
& 76.47
& 75.44
& 43.64
& 67.27
& 65.22
& \textbf{46.81} \\

DeepSeek-V4-Pro
& 61.22
& 69.12
& 71.93
& 56.36
& \textbf{74.55}
& 63.77
& 14.89 \\

Grok-4.1-Reasoning
& 51.02
& 63.24
& 70.18
& 43.64
& 65.45
& 60.87
& 12.77 \\

GLM-5.1
& 53.06
& 57.35
& 57.89
& 47.27
& 61.82
& 63.77
& 12.77 \\

Claude Sonnet 4.6
& 53.06
& 57.35
& 61.40
& 52.73
& 56.36
& 59.42
& 8.51 \\

GPT-5-mini
& 46.94
& 52.94
& 57.89
& 38.18
& 60.00
& 59.42
& 19.15 \\

Gemini-2.5-Lite
& 6.12
& 0.00
& 0.00
& 3.64
& 10.91
& 2.90
& 6.38 \\

Llama-3.3-70B
& 2.04
& 2.94
& 0.00
& 1.82
& 0.00
& 2.90
& 0.00 \\

\bottomrule
\end{tabular}
\end{table}

\clearpage

\section{Tools}

\label{app:appb}
\begin{table}[H]
\centering
\small
\begin{tabular}{p{0.13\textwidth} p{0.55\textwidth} p{0.24\textwidth}}
\hline
\textbf{Category} & \textbf{Tools} & \textbf{Use} \\
\hline
Search &
\texttt{search\_issues}, \texttt{search\_code}, \texttt{search\_pull\_requests}, \texttt{search\_repositories}, \texttt{search\_users} &
Open-text queries over GitHub's index \\
\hline
Issues &
\texttt{issue\_read} with methods: \texttt{get}, \texttt{get\_comments}, \texttt{get\_sub\_issues}, \texttt{get\_labels}; \texttt{list\_issues} (2) &
Fetch and list issue details \\
\hline
Pull requests &
\texttt{pull\_request\_read} with methods: \texttt{get}, \texttt{get\_diff}, \texttt{get\_status}, \texttt{get\_files}, \texttt{get\_review\_comments}, \texttt{get\_reviews}, \texttt{get\_comments}, \texttt{list\_pull\_requests} &
Fetch and list PR details \\
\hline
Commits &
\texttt{get\_commit}, \texttt{list\_commits}&
Fetch commit details and history \\
\hline
Branches &
\texttt{list\_branches} &
Enumerate branches \\
\hline
Tags &
\texttt{list\_tags}, \texttt{get\_tag}&
Enumerate / inspect tags \\
\hline
Releases &
\texttt{list\_releases}, \texttt{get\_latest\_release}, \texttt{get\_release\_by\_tag} &
Enumerate / inspect releases \\
\hline
Labels &
\texttt{list\_label}, \texttt{get\_label}&
Inspect issue labels \\
\hline
Files &
\texttt{get\_file\_contents}, \texttt{get\_repository\_tree}&
Read repository file contents and tree \\
\hline
Answer &
\texttt{submit\_answer}&
Final answer with confidence $\in \{\text{high}, \text{medium}, \text{low}\}$ and reasoning \\
\hline
\end{tabular}
\caption{Tool categories available to the agent for interacting with GitHub.}
\label{tab:github-tools}
\end{table}

\section{Construction Prompts}
\label{app:construction-prompts}

To support reproducibility, we release the complete prompts used for the
LLM-assisted stages of the \textsc{RepoRef} construction pipeline.
These include the prompts used for conversation segmentation, masking,
and filtering. The prompts are available in the accompanying code
repository under:

\begin{center}
\texttt{reporef-benchmark/reporef/config/prompts}
\end{center}

We provide the prompts in the repository rather than reproducing them
in full here to keep the appendix concise and ensure that the released
prompts remain directly aligned with the benchmark implementation.

\clearpage
\raggedbottom
\section{Refute Verifier}
\label{app:appc}

\subsection{Metadata Cue Candidate Refutation}
We use a metadata-cue verifier that compares the submitted target against expectations directly inferred from the conversation, such as target type, GitHub item author, and temporal constraints. For example, if in the anchor reference message the user says "the issue I opened" it can be assumed that the issue was created by that author. The verifier is refutational: it can flag inconsistencies between a candidate and the conversation, but it does not certify that an answer is correct.

Across models, metadata cues alone refute between one third and one half of wrong predictions. Refutability decreases with model strength; weaker models often choose targets that violate more metadata expectations, while stronger models tend to choose candidates that are more plausible but still incorrect by "softer" cues from the conversation. The most common refutation is date mismatch, accounting 20\% of wrong predictions in average across all models.

These findings highlight a current limitation of CoRG agents: they often submit targets that remain inconsistent with conversationally implied constraints. This motivates lightweight verification as a future direction.

\begin{table}[H]
\centering
\small
\label{tab:per-model-per-cue-400-no-title}
\setlength{\tabcolsep}{4pt}
\begin{tabular}{lrrrrrrr}
\toprule
\textbf{Model} & \textbf{Acc (B=10)} & \textbf{Refute (any)} & \textbf{State} & \textbf{Date} & \textbf{Author} & \textbf{Repo} & \textbf{Type} \\
\midrule
Gemini 3 Flash & 67.0\% & 36.9\% & \textbf{17.1\%} & 10.8\% & 6.5\% & 1.8\% & 6.3\% \\
DeepSeek V4 Pro & 60.2\% & 34.9\% & 9.9\% & \textbf{14.5\%} & 1.3\% & 2.0\% & 9.9\% \\
Grok 4.1 (reasoning) & 54.0\% & 43.1\% & 13.1\% & \textbf{20.0\%} & 5.0\% & 2.5\% & 10.6\% \\
GLM-5.1 & 52.0\% & 28.2\% & 7.7\% & \textbf{9.9\%} & 2.1\% & 2.8\% & 8.5\% \\
Claude Sonnet 4.6 & 51.2\% & 38.3\% & \textbf{17.5\%} & 16.9\% & 4.5\% & 2.6\% & 7.1\% \\
GPT-5 mini & 49.0\% & 44.7\% & 14.5\% & \textbf{22.3\%} & 5.0\% & 4.5\% & 6.7\% \\
Gemini 2.5 Flash Lite & 4.0\% & 63.4\% & 12.7\% & \textbf{26.8\%} & 8.5\% & 12.7\% & 25.4\% \\
Llama 3.3 70B & 1.5\% & 53.5\% & 19.2\% & \textbf{31.4\%} & 10.3\% & 7.3\% & 4.1\% \\
\bottomrule
\end{tabular}
\caption{Per-model refutation rates on the RepoRef-400 subset using only meta-data cues. Each cell is the percentage of the model's wrong answers refuted by a mismatch on that field under the lenient (any-verified-cue) rule; ``Refute (any)'' is the per-model union across the non-title cues.}
\end{table}

\subsection{Qualitative Refute Examples}
\label{sec:qualitative-refute}
\begin{table}[H]
\centering
\footnotesize
\setlength{\tabcolsep}{4pt}
\renewcommand{\arraystretch}{1.08}
\caption{Qualitative examples of metadata-cue refutations. \textbf{Chat-implied} is extracted from the masked conversation; A mismatch refutes the candidate.}
\label{tab:qualitative-refute}

\begin{tabularx}{\textwidth}{@{}p{0.85cm} >{\raggedright\arraybackslash}p{0.36\textwidth} >{\raggedright\arraybackslash}p{0.38\textwidth} >{\raggedright\arraybackslash}X@{}}
\toprule
Cue & Chat snippet & Refutation signal & How refuted \\
\midrule
\midrule
State &
\textbf{[user\_b]} \textbf{[REFERENCE]} should we wait for feedback on the PR I opened a few days ago and then backport it?\par
\textbf{[user\_b]} warnings in the beta seem not that bad\par
\ldots
&
\textbf{GT title:} \textit{[MRG+2] Pass include\_self=True to kneighbors\_graph}\par
\textbf{Chat-implied:} \texttt{open}\par
\textbf{Gemini picked:} \textit{[MRG+1] Isotonic regression duplicate fixes}\par
\textbf{Candidate value:} \texttt{merged}
&
Chat implies an open PR; candidate was already merged/closed.
\\

\midrule
Date &
\textbf{[user\_a]} its all do able, but unless someone takes it on.....\par
\textbf{[user\_e]} \textbf{[REFERENCE]} The multisampling missing should've been added in that PR from a couple weeks ago. But still, the way t...\par
\textbf{[user\_e]} Also as is you can only set the depth and back buffer format in your game constructor
&
\textbf{GT title:} \textit{[DesktopGL] General Fixes}\par
\textbf{Chat-implied:} \texttt{2017-02-01 -2017-02-25}\par
\textbf{Gemini picked:} \textit{DesktopGL was not using back buffer and depth buffer formats}\par
\textbf{Candidate value:} \texttt{2016-01-28}
&
Candidate was created outside the implied date window.
\\

\midrule
Author &
\textbf{[user\_b]} here\par
\textbf{[user\_c]} \textbf{[REFERENCE]} you can share it here in the issue I opened -\textgreater{}\par
\textbf{[user\_b]} [![Simulator Screen Shot Mar 11, 2017, 11.44.31 PM.png](https://files.gitter.im/patchthecode/JTAppleCalendar/CFRA/thumb/Simulat...\par
\ldots
&
\textbf{GT title:} \textit{Made a cool calendar? Post its image here. \#2}\par
\textbf{Chat-implied:} user\_c\par
\textbf{Gemini picked:} \textit{Reduce number of columns?}\par
\textbf{Candidate value:} user\_f
&
Chat speaker says they authored the artifact; candidate is by a different user.
\\

\midrule
Type &
\textbf{[user\_c]} @mrhelmut what happened?\par
\textbf{[user\_e]} \textbf{[REFERENCE]} @user\_d looks like a general refactor of some core application stuff in that commit from last week\par
\textbf{[user\_c]} I'll have to take a look see if its broken anything\par
\ldots
&
\textbf{GT title:} \textit{Merge pull request \#5468 from cra0zy/desktopglfixcentering}\par
\textbf{Chat-implied:} \texttt{commit}\par
\textbf{Gemini picked:} \textit{[DesktopGL] General Fixes}\par
\textbf{Candidate value:} \texttt{pr}
&
Chat implies a commit; candidate is a PR.
\\

\bottomrule
\end{tabularx}
\end{table}

\clearpage

\section{Discovery Latency}
\label{app:appd}
\begin{table}[h]
\centering\small
\caption{Discovery latency per model over the canonical~400 capability examples (budget $=$ 10). Each $\leq k$ column is the percent of runs where GT first appeared in (or before) the $k$-th tool call's result (Layers 1+2: search items or direct fetch). The rightmost column is the mean percent of tool calls per run that failed JSON-schema validation.}
\label{tab:discovery-latency}
\setlength{\tabcolsep}{4pt}
\begin{tabular}{l r rrrrrrrrrr r}
\toprule
\textbf{Model} & \textbf{Step$\leq$1} & \textbf{Step$\leq$2} & \textbf{Step$\leq$3} & \textbf{Step$\leq$4} & \textbf{Step$\leq$5} & \textbf{Step$\leq$6} & \textbf{Step$\leq$7} & \textbf{Step$\leq$8} & \textbf{Step$\leq$9} & \textbf{Step$\leq$10} \\
\midrule
gemini-3-flash  & 24.5 & 42.5 & 52.5 & 59.5 & 62.5 & 66.0 & 69.2 & 72.0 & 73.5 & \textbf{75.0} \\
grok-4-1-fast-reasoning  & 22.0 & 32.0 & 37.8 & 38.5 & 38.5 & 38.5 & 38.5 & 38.5 & 38.5 & \textbf{38.5} \\
gpt-5-mini  & 21.8 & 31.5 & 39.2 & 44.0 & 47.2 & 49.0 & 51.8 & 52.2 & 53.0 & \textbf{53.8} \\
claude-sonnet-4-6  & 21.0 & 32.2 & 39.2 & 42.8 & 44.8 & 47.0 & 48.5 & 49.0 & 49.2 & \textbf{49.2} \\
deepseek-v4-pro & 18.5 & 30.2 & 38.2 & 45.0 & 50.5 & 53.5 & 54.8 & 55.2 & 55.5 & \textbf{55.8} \\
glm-5-1 & 12.8 & 20.8 & 27.0 & 34.5 & 40.8 & 43.2 & 44.2 & 44.8 & 45.0 & \textbf{45.0} \\
gemini-2-5-flash-lite & 2.2 & 3.5 & 3.8 & 4.2 & 4.2 & 5.0 & 5.0 & 5.2 & 5.2 & \textbf{5.2} \\
llama-3-3-70b & 0.2 & 0.8 & 0.8 & 0.8 & 0.8 & 0.8 & 0.8 & 0.8 & 0.8 & \textbf{0.8} \\
\bottomrule
\end{tabular}

\end{table}
\clearpage

\section{Rates of zero-tool-call runs and tool-call errors by model}

\begin{table}[h]
\centering\small
\caption{Per-model count of zero-tool-call runs on the canonical~400 capability examples at budget~$=$~10. A zero-tool-call run is one where the model issued no tool calls before submitting an answer}
\label{tab:tool-call-counts-error-rate}
\begin{tabular}{lrrrr}
\toprule
\textbf{Model} & \textbf{Zero-tool-call} & \textbf{Tool error rate (\%)} \\
\midrule
gemini-2-5-flash-lite  & 65.5 & 2.2 \\
llama-3-3-70b & 5.8 & 72.9 \\
gpt-5-mini & 4.5 & 0.5 \\
claude-sonnet-4-6 & 7 & 0.2 \\
grok-4-1-fast-reasoning & 4 & 0.0 \\
deepseek-v4-pro & 0 & 1.9 \\
gemini-3-flash & 0 & 2.2 \\
glm-5-1  & 0 & 0.0 \\
\bottomrule
\end{tabular}

\end{table}

\clearpage

\section{Decomposing Search and Disambiguation Failures}

CoRG combines two fundamental capabilities: searching for the relevant artifact in a large external workspace, and identifying the intended artifact among plausible candidates. Since benchmark examples naturally require both abilities, overall accuracy alone does not reveal which stage is responsible for failures. To better localize errors, we decompose unsuccessful trajectories into two observable categories: (1) \emph{pre-surfacing failures}, where the gold artifact is never discovered during search, and (2) \emph{post-surfacing failures}, where the gold artifact is surfaced but the agent ultimately selects a different candidate. This decomposition provides a more fine-grained view of where current agents struggle and helps interpret the analyses that follow.

\begin{table*}[!ht]
\centering
\small
\setlength{\tabcolsep}{5pt}
\begin{tabular}{lrrrr}
\toprule
Model
& Acc. (\%)
& Discovery Failures (\%)
& Selection Failures (\%)
& $P(\text{correct} \mid \text{surfaced})$ (\%) \\
\midrule
DeepSeek-V4-Pro
& 60.25
& 86.8
& 13.2
& 90.7 \\

Grok-4.1
& 54.00
& 92.4
& 7.6
& 90.9 \\

GLM-5.1
& 52.00
& 91.1
& 8.9
& 90.6 \\

Claude Sonnet 4.6
& 51.25
& 89.2
& 10.8
& 89.3 \\

GPT-5-mini
& 49.00
& 79.9
& 20.1
& 80.9 \\

Claude Code Opus 4.7
& 63.25
& 80.0
& 20.0
& 89.3$^{\dagger}$ \\

Gemini-3-Flash
& 67.00
& 70.5
& 29.5
& 87.0 \\
\bottomrule
\end{tabular}
\caption{
Model-wise decomposition of unsuccessful runs into discovery and selection failures.
Discovery failures are runs in which the gold artifact never appears in the agent trajectory.
Selection failures are runs in which the gold artifact is surfaced, but the agent submits another artifact.
The final column reports the probability of selecting the correct artifact conditional on surfacing it.
$^{\dagger}$Approximate value.
}
\label{tab:failure-decomposition}
\end{table*}

\clearpage

\section{Answer Composition}
We additionally report trajectory diagnostics: invalid tool-use rate, hallucination rate, and information-gain rate. Invalid tool use covers malformed or unsupported tool calls; hallucination captures incorrect submitted URLs that do not resolve to valid GitHub targets; and information gain measures the fraction of tool calls that surface at least one previously unseen issue, pull request, or commit.
\vspace{\baselineskip}
\label{app:appe}
\begin{table}[h]
\centering\small

\label{tab:answer-composition}
\begin{tabular}{lrrrrrr}
\toprule
\textbf{Model} & \textbf{Tool Budget} & \textbf{Correct} & \textbf{\makecell{Wrong\\but real}} & \textbf{Hallucinated} & \textbf{Malformed} & \textbf{\makecell{No\\prediction}} \\
\midrule
gemini-3-flash & 1 & 22.2 & 52.8 & 20.8 & 2.0 & 2.2 \\
gemini-3-flash & 3  & 50.0 & 36.0 & 11.0 & 1.5 & 1.5 \\
gemini-3-flash & 6  & 64.8 & 27.2 & 6.5 & 1.0 & 0.5 \\
gemini-3-flash & 10  & 67.0 & 26.2 & 6.5 & 0.2 & 0.0 \\
deepseek-v4-pro & 10 & 60.0 & 37.8 & 2.0 & 0.2 & 0.0 \\
grok-4-1-fast-reasoning & 10 & 54.0 & 38.8 & 2.2 & 2.2 & 2.8 \\
glm-5-1 & 10 & 52.0 & 32.2 & 1.8 & 3.8 & 10.2 \\
claude-sonnet-4-6 & 10 & 51.2 & 39.0 & 3.8 & 2.8 & 3.2 \\
gpt-5-mini & 10 & 49.0 & 41.8 & 1.0 & 3.8 & 4.5 \\
gemini-2-5-flash-lite & 10 & 3.8 & 12.5 & 2.2 & 3.2 & 78.2 \\
llama-3-3-70b & 10 & 1.5 & 71.2 & 22.5 & 1.2 & 3.5 \\
\bottomrule
\end{tabular}
\caption{Per-model answer composition over the canonical~400 capability examples (budget $=$ 10). Values are percent of runs. \emph{Correct} matches GT; \emph{wrong but real} is a different artifact URL that exists on GitHub; \emph{hallucinated} is a URL form that returns 404/410/422 (invented artifact); \emph{malformed} is a predicted value that is not a parseable artifact URL; \emph{no prediction} is a run where the agent never produced a URL.}
\end{table}

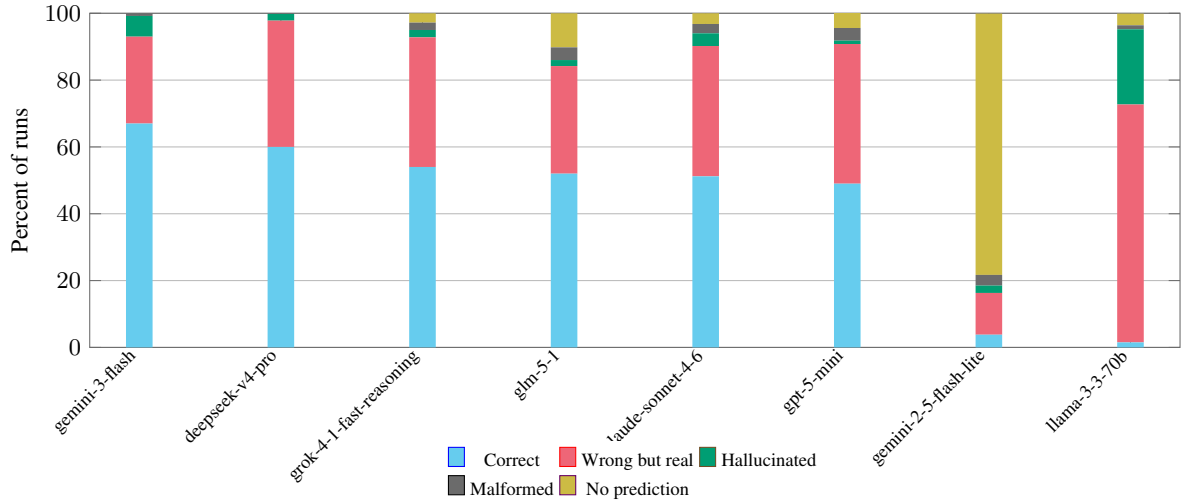
\begin{figure}[H]
\centering
\small
\label{fig:answer-composition}
\begin{tikzpicture}
\begin{axis}[
    ybar stacked,
    bar width=10pt,
    width=\columnwidth,
    height=6cm,
    ymin=0,
    ymax=100,
    ylabel={Percent of runs},
    symbolic x coords={
        gemini-3-flash,
        deepseek-v4-pro,
        grok-4-1-fast-reasoning,
        glm-5-1,
        claude-sonnet-4-6,
        gpt-5-mini,
        gemini-2-5-flash-lite,
        llama-3-3-70b
    },
    xtick=data,
    x tick label style={
        rotate=45,
        anchor=east,
        font=\scriptsize
    },
    ymajorgrids=true,
    grid style={Gray},
    enlarge x limits=0.05,
    legend style={
        at={(0.5,-0.28)},
        anchor=north,
        legend columns=3,
        draw=none,
        font=\scriptsize
    },
    axis line style={DarkGray},
    tick style={DarkGray}
]

\addplot+[fill=MintWave, draw=none] coordinates {
    (gemini-3-flash,67.0)
    (deepseek-v4-pro,60.0)
    (grok-4-1-fast-reasoning,54.0)
    (glm-5-1,52.0)
    (claude-sonnet-4-6,51.2)
    (gpt-5-mini,49.0)
    (gemini-2-5-flash-lite,3.8)
    (llama-3-3-70b,1.5)
};

\addplot+[fill=Pink, draw=none] coordinates {
    (gemini-3-flash,26.0)
    (deepseek-v4-pro,37.8)
    (grok-4-1-fast-reasoning,38.8)
    (glm-5-1,32.2)
    (claude-sonnet-4-6,39.0)
    (gpt-5-mini,41.8)
    (gemini-2-5-flash-lite,12.5)
    (llama-3-3-70b,71.2)
};

\addplot+[fill=DeepSpaceGreen, draw=none] coordinates {
    (gemini-3-flash,6.2)
    (deepseek-v4-pro,2.0)
    (grok-4-1-fast-reasoning,2.2)
    (glm-5-1,1.8)
    (claude-sonnet-4-6,3.8)
    (gpt-5-mini,1.0)
    (gemini-2-5-flash-lite,2.2)
    (llama-3-3-70b,22.5)
};

\addplot+[fill=DarkGray, draw=none] coordinates {
    (gemini-3-flash,0.8)
    (deepseek-v4-pro,0.2)
    (grok-4-1-fast-reasoning,2.2)
    (glm-5-1,3.8)
    (claude-sonnet-4-6,2.8)
    (gpt-5-mini,3.8)
    (gemini-2-5-flash-lite,3.2)
    (llama-3-3-70b,1.2)
};

\addplot+[fill=Mustard, draw=none] coordinates {
    (gemini-3-flash,0.0)
    (deepseek-v4-pro,0.0)
    (grok-4-1-fast-reasoning,2.8)
    (glm-5-1,10.2)
    (claude-sonnet-4-6,3.2)
    (gpt-5-mini,4.5)
    (gemini-2-5-flash-lite,78.2)
    (llama-3-3-70b,3.5)
};

\legend{Correct, Wrong but real, Hallucinated, Malformed, No prediction}

\end{axis}
\end{tikzpicture}
\caption{Per-model answer composition over the canonical~400 capability examples (budget $=10$). Values are percent of runs.}
\end{figure}

\end{document}